\documentstyle[12pt]{article}   
\begin{document}
\bibliographystyle{plain}
\hyphenation{mono-tonicity Mono-tonicity mono-tonic Mono-tonic Mo-notonicity
mo-notonicity monoto-nicity Monoto-nicity}
%%%%%%%%%%%%%%%%%%%%%%%%%%%%%%%%%%%%%%%%%%%%%%%%%%%%%%%%%%%%%%
%%%%%%%%%%%%%%%%%%%%%%%%%%%%%%%%%%%%%%%%%%%%%%%%%%%%%%%%%%%%%%
\newtheorem{theorem}{Theorem}
\newtheorem{corollary}{Corollary}
\newtheorem{lemma}{Lemma}
\newtheorem{exercise}{Exercise}
\newtheorem{claim}{Claim}
\newtheorem{remark}{Remark}
\newtheorem{definition}{Definition}
\newtheorem{example}{Example}
\newenvironment{notation}{\noindent\bf Notation:\em\penalty1000}{}
%%%%%%%%%%%%%%%%%%%%%%%%%%%%%%%%%%%%%%%%%%%%%%%%%%%%%%%%%%%%%%
\newcommand{\blackslug}{\mbox{\hskip 1pt \vrule width 4pt height 8pt 
depth 1.5pt \hskip 1pt}}
\newcommand{\QED}{\quad\blackslug\lower 8.5pt\null\par\noindent}
\newcommand{\proof}{\par\penalty-1000\vskip .5 pt\noindent{\bf Proof\/: }}
%%%%%%%%%%%%%%%%%%%%%%%%%%%%%%%%%%%%%%%%%%%%%%%%%%%%%%%%%%%%%%
%\newcommand{\SEP}{\makebox[0in]{\rule{.5mm}{4.5mm}}}
\newcommand{\ru}{\rule[-0.4mm]{.1mm}{3mm}}
\newcommand{\nni}{\ru\hspace{-3.5pt}}
\newcommand{\sni}{\ru\hspace{-1pt}}
\newcommand{\pre}{\hspace{0.28em}}
\newcommand{\post}{\hspace{0.1em}}
\newcommand{\NIm}{\pre\nni\sim}
\newcommand{\NI}{\mbox{$\: \nni\sim$}}
\newcommand{\notNIm}{\pre\nni\not\sim}
\newcommand{\notNI}{\mbox{ $\nni\not\sim$ }}
\newcommand{\NIVm}{\pre\nni\sim_V}
\newcommand{\NIV}{\mbox{ $\nni\sim_V$ }}
\newcommand{\notNIVm}{\pre\sni{\not\sim}_V\post}
\newcommand{\notNIV}{\mbox{ $\sni{\not\sim}_V$ }}
\newcommand{\NIWm}{\pre\nni\sim_W}
\newcommand{\NIW}{\mbox{ $\nni\sim_W$ }}
\newcommand{\NIWp}{\mbox{ $\nni\sim_{W'}$ }}
\newcommand{\notNIWm}{\pre\sni{\not\sim}_W\post}
\newcommand{\notNIW}{\mbox{ $\sni{\not\sim}_W$ }}
\newcommand{\eem}{\hspace{0.8mm}\rule[-1mm]{.1mm}{4mm}\hspace{-4pt}}
\newcommand{\EM}{\eem\equiv}
\newcommand{\notEM}{\eem\not\equiv}
\newcommand{\R}{\cal R}
\newcommand{\notR}{\not {\hspace{-1.5mm}{\cal R}}}
%%%%%%%%%%%%%%%%%%%%%%%%%%%%%%%%%%%%%%%%%%%%
\newcommand{\bK}{{\bf K}}
\newcommand{\bKp}{${\bf K}^p$}
\newcommand{\oK}{$\overline {\bf K}$}
\newcommand{\bM}{{\bf M}}
\newcommand{\bP}{{\bf P}}
\newcommand{\ga}{\mbox{$\alpha$}}
\newcommand{\gb}{\mbox{$\beta$}}
\newcommand{\gc}{\mbox{$\gamma$}}
\newcommand{\gd}{\mbox{$\delta$}}
\newcommand{\gep}{\mbox{$\varepsilon$}}
\newcommand{\gf}{\mbox{$\zeta$}}
\newcommand{\cA}{\mbox{${\cal A}$}}
\newcommand{\cB}{\mbox{${\cal B}$}}
\newcommand{\cC}{\mbox{${\cal C}$}}
\newcommand{\cE}{\mbox{${\cal E}$}}
\newcommand{\cF}{\mbox{${\cal F}$}}
\newcommand{\cK}{\mbox{${\cal K}$}}
\newcommand{\cL}{\mbox{${\cal L}$}}
\newcommand{\cM}{\mbox{${\cal M}$}}
\newcommand{\cR}{\mbox{${\cal R}$}}
\newcommand{\cS}{\mbox{${\cal S}$}}
\newcommand{\cT}{\mbox{${\cal T}$}}
\newcommand{\cU}{\mbox{${\cal U}$}}
\newcommand{\ab}{\mbox{\ga \NI \gb}}
\newcommand{\cd}{\mbox{\gc \NI \gd}}
\newcommand{\ef}{\mbox{\gep \NI \gf}}
\newcommand{\xe}{\mbox{$\xi$ \NI $\eta$}}  
\newcommand{\pht}{\mbox{$\varphi$ \NI $\theta$}}
\newcommand{\rt}{\mbox{$\rho$ \NI $\tau$}}
\newcommand{\Cn}{\mbox{${\cal C}n$}}
\newcommand{\CF}{\mbox{${\cal C}_{\cal F}$}}
\newcommand{\CG}{\mbox{${\cal C}_{\sim}$}}
\newcommand{\CW}{\mbox{${\cal C}_{W}$}}
\newcommand{\Pf}{\mbox{${\cal P}_{f}$}}
\newcommand{\leC}{\mbox{${\preceq_{\cC}}$}}
\newcommand{\leF}{\mbox{${\preceq_{\cF}}$}}
\newcommand{\lC}{\mbox{${\prec_{\cC}}$}}
\newcommand{\notlC}{\mbox{$\not \! \! \lC$}}
\newcommand{\lF}{\mbox{${\prec_{\cF}}$}}
\newcommand{\Z}{\mbox{$Z_{\cF}$}}
%%%%%%%%%%%%%%%%%%%%%%%%%%%%%%%%%%%%%%%%%%%%
\newcommand{\ra}{\rightarrow}
\newcommand{\Ra}{\Rightarrow}
\newcommand{\eqdef}{\stackrel{\rm def}{=}}
\newcommand{\absv}[1]{\mid #1 \mid}
\newcommand{\vstar}{\mbox{$V\sstar_{\infty}$}}
\newcommand{\sumstar}{\mbox{$\sum\sstar$}}
\newcommand{\tilh}{\mbox{$\tilde{h}$}}
\newcommand{\tilep}{\mbox{$\tilde{\varepsilon}$}}
\newcommand{\tilf}{\mbox{$\tilde{f}$}}
\newcommand{\gahat}{\mbox{$\hat{\ga}$}}
\newcommand{\gafalse}{\mbox{$\ga\NI{\bf false}$}}
\newcommand{\sstar}{^{*}}
\newcommand{\calR}{\mbox{${\cal R}\sstar$}}
\newcommand{\subseteqf}{\mbox{$\subseteq_{f}$}}

%%%%%%%%%%%%%%%%%%%%%%%%%%%%%%%%%%%%%%%%%%%%%%%%%%%%%%%%%%%%%%
\title{Another perspective on Default Reasoning
\thanks{
This work was 
partially supported 
by the Jean and Helene Alfassa fund for 
research in Artificial Intelligence}
}
\author{Daniel Lehmann \\ Institute of Computer Science, \\
Hebrew University, Jerusalem 91904 (Israel)
}
\date{}

\maketitle
\begin{abstract}
The lexicographic closure of any given finite set $D$ of normal defaults
is defined. A conditional assertion \mbox{$a \NIm b$} is in this lexicographic
closure if, given the defaults $D$ and the fact $a$, one would conclude
$b$. The lexicographic closure is essentially a rational extension of
$D$, and of its rational closure, defined in a previous paper.
It provides a logic of normal defaults that is different from the one
proposed by R. Reiter and that is rich enough not to require the consideration
of non-normal defaults.
A large number of examples are provided to show that the lexicographic
closure corresponds to the basic intuitions behind Reiter's logic of 
defaults.
\end{abstract}

\section{Plan of this paper}
\label{sec:plan}
Section~\ref{sec:intro} is a general introduction, describing the goal
of this paper, in relation with Reiter's Default Logic and the program
proposed in~\cite{LMAI:92} by Lehmann and Magidor.
Section~\ref{sec:nonmonreas} first discusses at length some general
principles of the logic of defaults, with many examples, and, then, 
puts this paper in perspective relatively to previous work.
Section~\ref{sec:what} sets the stage for this paper by describing
the intuitive meaning of default information and the formal representation
used in this paper for defaults. It singles out two different
possible interpretations for defaults: a {\em prototypical} and
a {\em presumptive} reading.
Section~\ref{sec:defvsmat} briefly discusses the relation between
defaults and material implications.
This paper proposes a meaning to any set $D$ of defaults.
This meaning is presented in a complex construction, that is
described in full only in Section~\ref{sec:full}.
The different aspects of this construction are presented
separately at first.
In Section~\ref{sec:single}, the meaning of a set consisting of a single
default will be studied. Reiter's proposal does not enable the use
of a default the antecedent of which is not known to hold.
The new perspective of this paper allows much more sophisticated ways of using 
default information.
In particular the default $(a : b)$ may be used to conclude that,
if $b$ is known to be false, then $a$ should be presumed to be false too.
Section~\ref{sec:seminormal} is a short digression on non-normal defaults.
It is shown that such defaults can never be understood if one 
requires that the closure of a set of defaults be rational.
Section~\ref{sec:Dsys} studies interacting normal defaults that
have the same rank (or strength). We propose that, in the case of 
contradictory defaults of the same rank, we try to satisfy as many
as possible. This proposal is in disagreement with D.~Poole's~\cite{Poole:88},
but in agreement with the Maximal Entropy approach of~\cite{GMP:90}.
It is shown that this idea guarantees rationality.
In Section~\ref{sec:full} a formal description of our complete 
proposal is given.
First, a model-theoretic construction is presented:
given a finite set $D$ of normal defaults, a modular model is defined and
the lexicographic closure of $D$ is the
rational consequence relation defined by this model.
Then, an equivalent characterization in terms of maxiconsistent sets is given.
Section~\ref{sec:exc} presents examples and the description of the answer 
provided by our proposal.
One of those shows how and why this proposal disagrees with 
the Maximal Entropy approach.
Section~\ref{sec:properties} is a concluding discussion.

\section{Introduction}
\label{sec:intro}
In~\cite{Reiter:80}, R. Reiter proposed a formal framework for 
Default Reasoning. Its focal point is the definition of an extension.
In~\cite{ReiterCri:83}, R. Reiter and G. Criscuolo found that, 
in this framework, one must consider non-normal defaults.
Non-normal defaults have, since then, been taken as the basic 
piece of {\em default} information by the logic programming community.
An alternative point of view is propounded here.
An answer is provided to the following question: 
given a set $D$ of normal defaults, what are the normal defaults that should
be considered as following from $D$, or entailed by $D$?
This answer provides a Logic of Defaults that does not suffer from the
problems of multiple extensions or the inabilty of Reiter's system to
cope satisfactorily with disjunctive information. There is no need to consider
non-normal defaults. 
In~\cite{LMAI:92}, M. Magidor and the author proposed, as their first
thesis [Thesis~1.1 there],
that the set of defaults entailed by any set $D$ be rational. 
This requirement is met.
The rational closure of a set $D$, defined there,
is not the set looked for, since it does not provide for inheritance
of generic properties to exceptional subclasses. 
In their second thesis [Thesis~5.25 in Section~5.9], they
proposed to look for some uniform way of constructing a rational superset
of the rational closure of a knowledge base.
The answer provided here, the lexicographic closure, is almost such a set,
and a simple variation meets the condition in full.
Independently, in~\cite{BCDLP:lexi}, Benferhat\&al. proposed a similar 
lexicographic construction based on an unspecified ordering of single 
defaults. When one applies their construction to the ordering on single 
defaults defined in~\cite{LMAI:92}, one obtains the lexicographic closure
presented in this paper.
Its computational complexity has been studied in~\cite{CL:comp} 
and~\cite{L:priv}: it is in
\mbox{$\Delta^{p}_{2}$} and is NP-hard and co-NP-hard.
The lexicographic closure is a syntactic construction in the sense 
of~\cite{Nebel:syntax}, i.e., it is sensitive to the presentation of the 
default information.

\section{Nonmonotonic Reasoning}
\label{sec:nonmonreas}
%Discuss Default Reasoning, normal and semi-normal defaults.
%Compare defaults statements and typicality statements.
%Explain the plan of the paper.
\subsection{The Rational Enterprise}
\label{subsec:rat}
We shall briefly summarize~\cite{KLMAI:89} and~\cite{LMAI:92}
and set up the stage.
This section was prepared in collaboration with David Makinson.
Some nonmonotonic inference relations are better behaved than others.
In particular, there are some simple closure conditions that appear
highly desirable: {\em reflexivity, left-logical-equivalence, 
right-weakening, and, or,} and {\em cautious monotonicity}.
The family of relations that satisfy those properties is closed under
intersection. Therefore, given a set \bK\ of ordered pairs 
\mbox{$(a , b)$} of formulas (which we shall write \mbox{$a \NIm b$} 
to remind us
that they are meant to be elements of an inference relation),
there is a natural and convincing way of defining a distinguished superset
of \bK\ that satisfies those conditions: simply put \bKp, 
called the preferential closure of \bK, to be the intersection of all 
supersets of \bK\ that satisfy the six conditions above.

However, there are other desirable ``closure'' (in a broad sense) properties
that are much more difficult to deal with.
{\em Rational monotonicity} defines a family of relations that is not closed
under intersection.
Other desirable conditions appear to be incapable of a purely formal
expression, but may be conveyed intuitively and are illustrable
by examples. Because of their informal nature, their identification is
not cut and dried, but four seem to be of particular interest:
\begin{enumerate}
\item \label{typ} the presumption of typicality,
\item \label{ind} the presumption of independence, 
\item \label{prio} priority to typicality, and
\item \label{spec} respect for specificity.
\end{enumerate}
There may be other desirable properties.

(\ref{typ}) The {\em presumption of typicality} begins where rational
monotonicity leaves off. Suppose \mbox{$ p \NIm x \in \bK$}.
By rational monotonicity, the {\em closure} of \bK, $\bK^{+}$, will contain
either \mbox{$p \wedge q \NIm x$} or \mbox{$p \NIm \neg q$}.
But which? No guideline is given. The presumption of typicality
(it may as well be called ``a weak presumption of monotonicity'')
tells us that, in the absence of a convincing reason to accept
the latter, we should prefer the former. 
\begin{example}
{\rm 
if \bK\ has \mbox{$p \NIm x$} as its sole element,
there is no apparent reason why the relation  \mbox{$\bK^{+} \supseteq \bK$}
that we regard as ``generated'' by \bK\ should contain \mbox{$p \NIm \neg q$}.
Hence, it should contain \mbox{$p \wedge q \NIm x$}.
{\em Note}: In this and all examples, 
$p , q , r \ldots x , y , z$ are understood to be 
{\em distinct} atomic formulas, i.e., propositional variables.
}
\end{example}

(\ref{ind}) The {\em presumption of independence} is a
sharpening of the presumption of typicality, and thus a stronger presumption
of monotonicity. For,
even if typicality is lost with respect to one consequent, we may still
presume typicality with respect to another, ``unless there is reason
to the contrary''.
\begin{example}
\label{indep}
{\rm
Suppose \mbox{$\bK = \{p \NIm x , p \NIm \neg q\}$}.
Presumption of typicality cannot be used to support \mbox{$p \wedge q \NIm x$},
since $\bK^{+}$ is known to contain \mbox{$p \NIm \neg q$}.
Presumption of independence tells us we should expect $x$ to be independent
of $q$, and therefore unaffected by the truth of $q$.
Therefore, it tells us, we should accept \mbox{$p \wedge q \NIm x$}.
}
\end{example}
\begin{example}
\label{indep2}
{\rm
Suppose \mbox{$\bK = \{p \NIm x , p \wedge q \NIm \neg x, p \NIm y\}$}.
Notice that \mbox{$p \NIm \neg q$} is in \bKp, the preferential closure
of \bK, and, therefore, the presumption of typicality cannot convince us
to accept \mbox{$p \wedge q \NIm y$}.
But, we should presume that $x$ is independent from $y$, as there is no reason
to think otherwise, and put \mbox{$p \wedge q \NIm y$} in the desired
consequence relation \mbox{$\bK^{+} \supseteq \bK$}.
}
\end{example}

{\em Remark:} The two conditions above may be interpreted 
as related to and strengthening the condition of rational monotonicity.
The difference between rational monotonicity, on one hand, and the
presumptions of typicality and independence is subtle, 
and may be easily overlooked. 
Rational monotonicity is a constraint on the product 
\mbox{$\bK^{+} \supseteq \bK$}, whereas presumptions of typicality and
independence are
best understood as rough and partial guides to the construction of a
desirable $\bK^{+}$.

(\ref{prio}) {\em Priority to typicality} tells us that, 
in a situation of clash
between two inferences, one of them based on the presumption of typicality,
the other one based on the presumption of independence, then 
we should prefer the former. Two examples are provided now.
\begin{example}
\label{strict}
{\rm
Suppose \mbox{$\bK = \{p \NIm x , p \wedge q \NIm \neg x\}$}.
The presumption of typicality offers \mbox{$p \wedge q \wedge r \NIm \neg x$},
since there is no compelling reason to accept \mbox{$p \wedge q \NIm \neg r$}.
The presumption of independence offers both 
\mbox{$p \wedge q \wedge r \NIm \neg x$} and
\mbox{$p \wedge q \wedge r \NIm x$}.
It clearly would not be justified to draw both conclusions.
Priority to typicality, tells us to prefer the former.
}
\end{example}
\begin{example}
\label{defeasible}
{\rm
Suppose \mbox{$\bK = \{p \NIm x , {\bf true} \NIm q , q \NIm \neg x\}$}.
The presumption of independence, acting on the last assertion of \bK, 
offers \mbox{$q \wedge p \NIm \neg x$}.
This is in conflict with
\mbox{$p \wedge q \NIm x$} that is offerred by presumption of typicality, from
the first assertion.
Priority to typicality says we should prefer the latter conclusion.
}
\end{example}

(\ref{spec}) {\em Respect for specificity} tells us that, in case of clash
between two presumptions, one of them based on an assertion with a more
specific antecedent than the other,
we should prefer the conclusion based on the more specific antecedent.
This principle is generally accepted and has been discussed in the literature.
It is somewhat difficult to formalize: what does ``based on'' mean?
It is closely related to the priority to typicality principle described
just above, but the exact relationship between those two principles
still needs clarification.
In examples~\ref{strict} and~\ref{defeasible}, the priority given
to typicality achieves precisely the respect for specificity we are looking 
for.
In example~\ref{strict}, we prefer to use \mbox{$p \wedge q \NIm \neg x$}
to \mbox{$p \NIm x$} also because $p \wedge q$ is strictly more specific
than $p$, i.e., \mbox{$p \wedge q \models p$}.
In example~\ref{defeasible}, we prefer to use \mbox{$p \NIm x$}
to \mbox{$q \NIm \neg x$} also because $p$ is defeasibly more specific
than $q$, since, from \mbox{${\bf true} \NIm q$}, we shall conclude
\mbox{$p \NIm q$} by presumption of typicality, or, preferably, presumption
of independence. Another, more technical, reason to view $p$ as more specific 
than $q$ is that the rank (the definition found in~\cite{LMAI:92} is explained
at the end of~\ref{subsec:introfull}) of $p$ is strictly greater than the 
rank of $q$.

Of course, along with the above principles, one should also not forget
{\em avoidance of junk}: the desired \mbox{$\bK^{+} \supseteq \bK$}
should avoid gratuitous additions (otherwise, e.g. the total relation
would do). In other words, $\bK^{+}$ should be, in some sense,
``least'' among the supersets of \bK\ satisfying the desired conditions.
``Least'' should certainly imply minimal in the set-theoretic sense,
i.e., no strict subset is acceptable,
but cannot mean ``included in any acceptable superset'', since our family
is not closed under intersection.

In~\cite{LMAI:92}, a construction is given, that, given any (finite) set \bK\
of pairs \mbox{$a \NIm b$} provides a rational extension \oK\ such that
\mbox{$\bK \subseteq \bK^{p} \subseteq \overline {\bK} = 
{\overline \bK}^{p} = \overline{\overline \bK}$}
that behaves well so far as the presumption of typicality and the respect
for specificity are concerned. However it does not pay much heed to the
presumption of independence. For example it does not legitimize the
conclusion \mbox{$p \wedge q \NIm y$} in example~\ref{indep2} above.
The purpose of the present paper is to propose a different construction
that performs better in this last respect, whilst not losing satisfaction
of the other formal and informal properties.
\subsection{Related Work}
R.~Reiter's~\cite{Reiter:80} was certainly one of the most influential
papers in the field of knowledge representation. 
It proposed a way of dealing with {\em default} information.
In short, it proposed to represent such information as normal defaults
and to define the meaning of a set of normal defaults as the set
of extensions it provides to any set of sentences.
In a follow-up paper~\cite{ReiterCri:83}, R.~Reiter and G.~Criscuolo
remarked that, in many instances, the simple-minded formalization of
situations involving more than one (normal) default was not adequate: 
the extension semantics enforced some unexpected and undesirable 
consequences. 
They proposed to cure this problem by considering an extended
class of defaults: semi-normal defaults.

In this paper, a different perspective on default reasoning is proposed.
Normal defaults are considered and sets of normal defaults are given
a meaning that is different from the one proposed in~\cite{Reiter:80}.
With this meaning, the interactions between defaults are as expected
and the consideration of non-normal defaults is superflous.
This perspective is in line with the first thesis of~\cite{LMAI:92},
that requires a set of defaults to define a rational consequence relation.
It is also almost in line with the second thesis of the same paper, 
that requires
a set of defaults to define a consequence relation that extends the rational
closure of the set of defaults, and a straightforward variation will
be shown to extend rational closure.
This goal of implementing Reiter's program, but with different
techniques, is similar to David Poole's~\cite{Poole:88}.
The present paper also shares some technical insights with Poole's. 
It may be considered also as a close relative of the maximal entropy
approach of~\cite{GMP:90,GMP:93}, but the semantics proposed here is different
from the one obtained from maximal entropy considerations.
This paper is a descendant of~\cite{LehTech:92}.
The main ideas of the lexicographic construction proposed in this paper
have been, independently, proposed in~\cite{BCDLP:lexi}. There, the 
initial ordering of single defaults was left for the user to chose.
A specific ordering of single defaults is used here.

\section{What is default information?}
\label{sec:what}
Default information is information about the way things usually are.
The paradigmatical example of such information,
that has been used by most researchers in the field, is 
{\em birds fly}. 
Syntactically, a default is a pair of propositions that will be written as
\mbox{$(a:b)$}, where $a$ and $b$ are formulas (of a propositional calculus
for this paper). Remember that only normal defaults are considered,
so that \mbox{$(a:b)$} is our notation for Reiter's
\mbox{${a : b} \over {b}$}.
The default \mbox{$({\bf true}:b)$} will written as \mbox{$( : b)$}.
Given a set $D$ of defaults representing some background information about 
the way things typically behave and a formula $a$ representing our
knowledge of the situation at hand, we shall ask what formulas should
be accepted as presumably true.
The meaning of a set of defaults $D$ will therefore be understood as
the set of pairs (conditional assertions) 
\mbox{$ c \NIm d $} it {\em entails},
i.e., for which $d$ should
be presumed to be true if $c$ is our knowledge about the specific situation,
i.e. represents the conjunction of the facts we know to be true.
It is probably reasonable to expect that the conditional assertion
\mbox{$ c \NIm d $} be entailed by a set $D$ containing the default
\mbox{$( c : d )$}, but this will be discussed in the sequel.
Notice, that we may, as well, consider that a set of normal defaults
entails a set a defaults, confusing ``snake'' (\NI) and colon (:).

The meaning of defaults is a delicate affair and it will be discussed
in depth now.
In~\cite{ReiterCri:83}, a {\em prototypical} reading is proposed: 
{\em birds fly}
being understood as {\em typical birds fly}. But, there is another possible
reading: {\em birds are presumed to fly unless there is evidence to the 
contrary}. This second reading will be called the {\em presumptive} reading.
The conclusions of this paper may be summarized in three sentences.
The two readings above are {\em almost} equivalent when isolated defaults are 
concerned,
they are {\em not} when sets of defaults are concerned. 
The rational closure construction of~\cite{LMAI:92} is the correct
formalization of the prototypical reading. The presumptive 
reading is the one
intended by Default Logic and its formalization is the topic of this paper.
The distinction between the two readings will be explained with an example.
This example is formally equivalent to the {\em Swedes} example described 
informally in~\cite[page 4]{LMAI:92}.
In this example, as in all other examples of this paper, the formulas
appearing in the defaults will be represented by meaningless letters
and not, as customary in the field, by meaningful sentences.
The remark that logic, the study of deductive processes, may
be concerned only with the form of the propositions, and not
with their meaning, dates back to Aristotle.
The use of semantically loaded formulas and the import of the reader's
knowledge of the world may only hamper the study of the formal 
properties of nonmonotonic deduction (that should perhaps be called induction).
When a given example is formally isomorphic to some well known folklore
example (or at least to some possible formalization of it), it will be 
pointed out.
\begin{example}[Swedes]
\label{ex:swedes}
{\rm Let $p$ and $q$ be different propositional variables.
Let $D$ be the set of two defaults: \mbox{$\{( :p) , ( :q) \}$}.
Accepting $D$ means that we believe, by default, that $p$ is true, and also
that $q$ is true.
Following the prototypical reading, then, 
{\em typically $p$ is true} and 
{\em typically $q$ is true}.
Following the presumptive reading {\em $p$ is presumed to be true unless 
there is evidence to the contrary} and {\em $q$ is presumed to be true unless 
there is evidence to the contrary}.
Suppose now that we have the information that \mbox{$\neg p \vee \neg q$} 
is true,
i.e., at least one of $p$ or $q$ is false.

Using the prototypical reading, we shall conclude that the situation at hand
is {\em not} typical. In such a case none of our two defaults is applicable:
{\em typically $p$ is true, but this is not a typical situation}, and therefore
we shall not conclude, even by default (i.e. defeasibly) that 
\mbox{$p \vee q$} holds true.

Using the presumptive reading, on the contrary, we shall conclude that
\mbox{$p \vee q$} should be presumed to be true unless there is evidence to 
the contrary, and since there
is no evidence of this sort, it should be presumed to be true.
We should therefore presume that exactly one of $p$ and $q$ holds.
}
\end{example}

\section{Default vs. Material Implication}
\label{sec:defvsmat}
A very natural feeling is that the meaning of any single default
\mbox{$( a : b )$} should be closely related to the meaning of the material
implication \mbox{$a \ra b$}. 
This last formula will be called the {\em material
counterpart} of the default \mbox{$( a : b )$}.
Similarly the meaning of a set of defaults $D$
should be related to the meaning of the set of its material counterparts.

It turns out that, both in the rational closure construction of~\cite{LMAI:92}
and in the construction proposed in this paper, the meaning of a default
\mbox{$( a : b )$} (that is an element of the set $D$ of defaults accepted
by a reasoner)
in the presence of knowledge $c$, either its material
counterpart \mbox{$a \ra b$} or void 
(i.e. equivalent to a tautology: {\bf true}).
Both constructions may therefore be described by pinpointing, given specific
information $c$, which of the defaults of $D$ are meaningful.
If $D_{c}$ is this set and $M_{c}$ the set of material counterparts
of $D_{c}$, then $d$ should be presumed true iff
$d$ is a logical consequence of \mbox{$c \cup M_{c}$}, i.e.,
\mbox{$M_{c} , c \models d$}.
This semantics fits well into the {\em implicit content} framework proposed
in~\cite{Stal:92}.

\section{Single Defaults}
\label{sec:single}
%Default statement =Typicality judgement
%Almost Rational Closure is our answer.
%Full description of the rational closure of (a:b).
Let $D$ be the singleton set \mbox{$\{ ( a : b ) \}$},
where $a$ and $b$ are arbitrary formulas.
We propose the following meaning to $D$:
\begin{itemize}
\item if the information at hand $c$ is consistent with \mbox{$a \ra b$},
i.e., \mbox{$c \not \models a \wedge \neg b$}
then the default is meaningful and $d$ is presumed iff 
\mbox{$ c , a \ra b \models d$}
\item otherwise, the default is meaningless and $d$ is presumed iff
\mbox{$c \models d$}.
\end{itemize}

An equivalent, more model-theoretic description, of the consequence
relation determined by $D$ is the following:
the rational consequence relation that is defined by the
modular model in which the propositional models are ranked in two levels:
on the bottom level (the more normal one) all models that satisfy
the material implication $a \ra b$, on the top level all other models.

This is the most natural understanding of the default information
{\em if $a$ is true then $b$ is presumably true}, 
and completely in line with D.~Poole's~\cite{Poole:88} treatment of
defaults.
Notice, though, that it does not always agree with Reiter's treatment and
only almost agrees with rational closure.
If the information at hand $c$ logically implies $a$, then the
perspective proposed here agrees with Reiter's: $d$ is presumed to be true iff
$d$ is an element of the unique extension of \mbox{$(D , \{c\})$}.

To see the difference with Reiter's treatment, suppose $a$ and $b$ are
different propositional variables and consider $c$ to be $\neg b$.
The perspective defended here will support the claim that 
\mbox{$\neg a$} should be presumed to be true, i.e., $a$ should be
presumed to be false.
For Reiter, on the contrary, there is a unique extension: \mbox{$\Cn(\neg b)$}
($\Cn$ is the logical consequence operator) 
and therefore we should not presume 
anything about $a$.
Similarly if $c$ is {\bf true}, the present perspective will support
\mbox{$a \ra b$}, whereas Reiter will not.

The comparison with rational closure is more subtle.
Our perspective agrees with rational closure except when
\mbox{$a \models \neg b$}.
This is quite an out of the ordinary situation: $a$ is logically equivalent
to something of the form
\mbox{$\neg b \wedge e$}, and the default is of the form
{\em if $b$ is false and $e$ is true, then assume $b$ is true}.
Such a default will probably never be used in practice,
but its consideration is nevertheless enlightening.
In such a situation, the present perspective claims that the meaning of the
default \mbox{$(\neg b \wedge e : b)$} is that all models that
satisfy \mbox{$\neg b \wedge e$} are on the top level.
In other terms, if 
\mbox{$c \not \models \neg b \wedge e$}, the default is meaningful
and means \mbox{$\neg b \wedge e \ra b$}, which is logically 
equivalent to \mbox{$e \ra b$}, but if \mbox{$c \models \neg b \wedge e$},
then the default is meaningless.
The treatment of this last case is well in line with the presumptive
reading: if $b$ is known to be false, then $b$ should not be assumed
to be true.
If we look at the way rational closure deals with this case, we see
that it agrees with the present perspective in the first case (i.e.,
if \mbox{$c \not \models \neg b \wedge e$}) but disagrees with it
in the second case. Rational closure accepts any conclusion from the
information that $e$ is true and $b$ is false.
This is in line with the prototypical reading of the default:
{\em if $e$ is true and $b$ is false, then typically $b$ is true}
may only mean that it is inconsistent for \mbox{$e \wedge \neg b$} to be true 
and therefore one should conclude anything when this happens.

The new perspective does not always support each member of the rational 
closure,
but the reader may check that the solution it supports is always rational
(in the technical sense of~\cite{LMAI:92}). A proof of this, in a more general
setting, will be given in Section~\ref{sec:full}.
How come our proposal is different from rational closure, that seemed
to be the only reasonable one?
Let \bK\ be the conditional 
knowledge base containing the single assertion 
\mbox{$e \wedge \neg b$ \NI $b$}.
The rational relation proposed here in place of the rational closure
does not contain \mbox{$e \wedge \neg b$ \NI $b$}.
It is not an extension of \bK\ and therefore does not satisfy Thesis~5.25:
``The set of assertions entailed by any set of assertions \bK\ is a
rational superset of the rational closure of \bK'' 
of~\cite{LMAI:92}.
This departure from Thesis~5.25 is not central to our proposal and a slight
variant of it would satisfy Thesis~5.25 by treating differently only those 
useless defaults discussed just above. This variant does not seem to fully fit
the presumptive reading of defaults, though.
If we denote by $\bK^{l}$ the (lexicographic) construction proposed 
in this paper, the variant we have in mind may be defined in the following way:
accept \mbox{$a \NIm b$} iff either $a$ has a rank for \bK\ and 
\mbox{$a \NIm b \in \bK^{l}$}, or $a$ has no rank.
This variant gives a superset (sometimes strict) of $\bK^{l}$, 
that is
also a superset (sometimes strict) of the rational closure \oK.
\section{Seminormal defaults}
\label{sec:seminormal}
This paper will show that, if one accepts a semantics that
is different from Reiter's, the reasons that compelled him to introduce
non-normal defaults disappear, and one may restrict oneself to normal defaults.
The reader may well ask whether one would not like to consider, anyway,
a more general form of defaults: the semi-normal defaults.
A semi-normal default \mbox{${a : e \wedge b}\over{b}$} means that 
{\em if $a$ is known to be true and there is no evidence that 
\mbox{$e \wedge b$} is false,
then $b$ should be presumed to be true}.
Let $a$ be a tautology, i.e., {\bf true} and $e$ and $b$ be different
propositional variables ($q$ and $p$ respectively).
Suppose we accept
the semi-normal default \mbox{${ : q \wedge p}\over{p}$}.
There is general agreement about the following points:
\begin{itemize}
\item if the information at hand $c$ is a tautology, 
i.e., we have no specific information, we should presume that $p$ is true,
since there is no evidence that \mbox{$q \wedge p$} does not hold,
\item if $c$ is $\neg q$, i.e., we know for sure that $q$ does not hold,
we should not use the default information and therefore we should {\em not}
presume $p$,
\item but, if we have no specific information, we should {\em not}
presume that $q$ holds (why should we?).
\end{itemize}
The three points above provide a counter-example to the rule of
Rational Monotonicity of~\cite{LMAI:92}:
we accept \mbox{${\bf true} \NIm p$}, but neither 
\mbox{${\bf true} \NIm \neg \neg q$}, nor \mbox{$\neg q \NIm p$}.
Even the simplest isolated non-normal default cannot be given 
a rational interpretation.
This remark is very important in view of the fact that the efforts
to harness Logic Programming to Nonmonotonic Reasoning take as their
basic component rules of the form
\[
a \leftarrow b \: , \neg c
\]
meaning {\em conclude $a$ if
$b$ has been concluded and $c$ cannot be concluded}.
This is essentially equivalent to considering the semi-normal default
\[ {b : a \wedge \neg c}\over{a},\] 
or to considering the not even semi-normal default
\[ {b : \neg c} \over {a} \]
and will lead to a consequence relation that is not rational.
All we have shown here is that non-normal defaults or the logic programming
approach to nonmonotonic reasoning are incompatible with the property of
rational monotonicity, which is central to this and previous papers.
\section{Competing but equal defaults}
\label{sec:Dsys}
%D-systems. Rationality. Musicians example.
After dealing, in Section~\ref{sec:single}, with single defaults,
we shall treat now the more interesting case of a set of interacting
normal defaults.
In general, given a set of defaults $D$, this set defines a ranking 
of the defaults, as explained in~\cite{LMAI:92}.
This ranking will be described in full in Section~\ref{subsec:introfull}.
The ranking of a default \mbox{$(a : b)$} relative to $D$ depends only on 
its antecedent $a$ and, as we shall see in Section~\ref{sec:exc},
defaults of higher ranking (they correspond to exceptions) should be 
considered stronger than those of lower ranking.
In this Section, we shall deal with the case all defaults have the same
rank, i.e., all defaults are equal in strength and none of them
correspond to an exception.
This happens only when all elements of $D$ have rank zero,
as will be clear in Section~\ref{subsec:introfull}.
It is clear that, when considering such defaults, we should always
assume that as many defaults as possible are satisfied (i.e. not violated).
We should therefore always prefer violating a smaller set of defaults
to violating a larger one. One may hesitate about the meaning
to be given to ``smaller'': set inclusion or smaller size.
The main conclusion of our considerations will be that sets of
defaults should (for rationality's sake) be compared by their size, 
not by set inclusion.

We choose an example isomorphic to the musicians example 
of~\cite[Section 4.4]{Gin:86}, but we shall first ask about it
questions that are different from those asked traditionally.
\begin{example}[Musicians]
{\rm 
Let $p$, $q$ and $r$ be different propositional variables.
Let $D$ contain the three following defaults:
\mbox{$\{( : p) , ( : q) , ( : r)\}$}.
In other words, $p$, $q$ and $r$ are assumed to hold by default.
If we learn that \mbox{$c \eqdef \neg p \wedge \neg q \vee \neg r$} holds,
i.e., that either both $p$ and $q$ are false, contrary to expectation,
or, also contrary to expectation, $r$ does not hold, what should we
assume?
Should one of the two possibilities (\mbox{$\neg p \wedge \neg q$}) and
\mbox{$\neg r$} be assumed more likely than the other one?

In~\cite{Poole:88}, D.~Poole claims we should not.
He claims there are two different maximal subsets of the material counterpart
of $D$ consistent with $c$ (two bases for $c$): \mbox{$\{p , q\}$}
and \mbox{$\{r\}$} and
he proposes that we presume true only those formulas that are both
in \mbox{$\Cn(r , c)$} and in \mbox{$\Cn(p , q , c)$}.
In particular, we should not presume $p$ to be true.
But, we should presume \mbox{$p \leftrightarrow q$} to hold.
Also, if we learn that \mbox{$c \wedge \neg p$} holds,
we should presume true only those formulas that are both in
\mbox{$\Cn(r , c , \neg p) = \Cn(r , c)$} and in 
\mbox{$\Cn(q , c , \neg p)$}. In particular, we should not
presume the truth of \mbox{$p \leftrightarrow q$}.
D.~Poole's proposal, therefore, does not satisfy the Rational
Monotonicity principle.

Guided by Thesis~1.1 of~\cite{LMAI:92}, that requires Rational Monotonicity,
a slight modification of Poole's ideas will be put forward now.
This modification is also supported by the Maximal Entropy approach 
of~\cite{GMP:90}.
The two bases above should not be considered equivalently plausible.
The larger one, which contains two defaults should be considered more 
plausible than the one containing only a single default.
In other terms, situations that violate two defaults should be considered
less plausible than those that violate only one default.
Here is a model-theoretic description.
We shall consider the (propositional) models of our language, and rank
them by the number of defaults of $D$ they violate.
A model $m$ violates a default \mbox{$a \NIm b$} iff it does not satisfy
the material implication \mbox{$a \ra b$}, i.e., iff 
\mbox{$m \models a \wedge \neg b$}.
The most normal models are those that violate no default of $D$:
they constitute the bottom level (zero) of our modular model.
Slightly less normal are those models that violate one single default:
they constitute level one of our model.
In general, level $i$ is constituted by all models that violate exactly
$i$ members of $D$. The nonmonotonic consequence relation defined
by this model is the one defined by $D$. It is rational, since the model
described is ranked and consequence relations
defined by modular models are rational (Lemma 3.9 of~\cite{LMAI:92}).
Coming back to the Musicians example: if we learn that 
\mbox{$c \eqdef \neg p \wedge \neg q \vee \neg r$} holds, \mbox{$\neg r$}
should be presumed true. We should therefore presume $p$ to be true.
}
\end{example}

We provided, just above, a model-theoretic description of our proposal.
An equivalent description in terms of ``bases'', in the spirit 
of~\cite{Poole:88}, is provided now. The same bases were also considered 
in~\cite{BCDLP:lexi}.
Let \mbox{$E = \{e_{i}\}$} be a finite set of formulas.
Let $c$ be a formula.
\begin{definition}
\label{def:maxbase}
A subset $F$ of $E$ is said to be a maxbase for $c$ iff $c$ is consistent
with $F$ and there is no subset $F^{\prime}$ of $E$,
\mbox{$\mid F^{\prime} \mid > \mid F \mid$}
that is consistent with $c$.
\end{definition}
\begin{theorem}
\label{the:rat}
Let $D$ be a set of defaults such that all elements of $D$ have rank zero
(with respect to $D$). Let $E$ be the set of material implications
corresponding to the defaults of $D$.
The consequence relation defined by the model-theoretic description above
is characterized by:
\begin{equation}
\label{eq1}
a \NIm b {\rm \ iff \ for \ every \ maxbase} \ F \ of \ E {\rm \ for} \ a \ \ 
F , a \models b.
\end{equation}.
\end{theorem}
Theorem~\ref{the:rat} implies that 
the relation defined by Equation~(\ref{eq1}) is rational.
\proof
First, some remarks.
Let $n$ be the size of the set $D$.
\begin{enumerate}
\item \label{aa}
If $a$ is satisfied by some model of level $i$ 
(\mbox{$0 \leq i \leq n$}),
then, there is a maxbase for $a$, and all maxbases for $a$ are of size
larger or equal to $n - i$.
\item \label{bb}
If $F$ is a maxbase of size $k$ (\mbox{$0 \leq k \leq n$}) for $a$,
then, there is a model of level $n - k$ that satisfies $a$, 
and no model of lower level satisfies $a$.
\item \label{cc}
There is no maxbase for $a$ iff $a$ is a logical contradiction.
\end{enumerate}
Suppose \mbox{$a \NIm b$}.
If no model satisfies $a$, $a$ is a logical contradiction and
\mbox{$ X , a \models b$} for any $X$.
Suppose, then, that $a$ is satisfied by some model of level $i$, 
but by no model of lower level. Any model of level $i$ that satisfies
$a$, satisfies $b$, by hypothesis. 
Let $F$ be a maxbase for $a$.
By remark~(\ref{bb}), \mbox{$ n \: - \mid F \mid \: = \: i$}.
Any model that satisfies $F$ is obviously of level less or equal to 
\mbox{$n \: - \mid F \mid \: = \: i$}.
Any model that satisfies $F$ and $a$ is therefore of level $i$
and satisfies $b$, by hypothesis. 
We conclude that \mbox{$F , a \models b$}.

Suppose, now, that for any maxbase $F$ for $a$ we have
\mbox{$F , a \models b$}.
If there is no maxbase for $a$, then, by remark~(\ref{cc}), 
$a$ is a logical contradiction and \mbox{$a \NIm b$}.
Suppose, then, the maxbases for $a$ are of size $k$.
There is, by remark~(\ref{bb}),  
a model of $a$ of level $n - k$, and there is no model
of level less than $n - k$ that satisfies $a$.
We must show that any model of $a$ of level $n - k$ satisfies $b$.
Let $m$ be such a model. Since $m$ violates $n - k$ defaults, it
satisfies a set $M$ of $k$ defaults. But $M$ is consistent with $a$,
since $m \models a$. The size of $M$ is the size of the maxbases for
$a$, therefore $M$ is a maxbase for $a$ and,
since \mbox{$M , a \models b$}, we conclude that \mbox{$m \models b$}.
\QED
\begin{example}[Musicians, continued]
{\rm
We shall now describe our solution to the questions traditionally 
asked about the musicians' example and generally used to demonstrate
that counterfactuals do not satisfy Rational Monotonicity.
Suppose our specific information is 
\mbox{$c \eqdef p \wedge \neg r \vee \neg p \wedge r$}.
There are two maxbases: \mbox{\{$p , q$\}} and \mbox{$\{q , r\}$}.
We shall therefore presume that $q$ holds and we shall {\em not} presume that 
\mbox{$d \eqdef q \wedge r \vee \neg q \wedge \neg r$} holds.
This is the common wisdom and the present
perspective subscribes to it.

Suppose now that our specific information is \mbox{$c \wedge \neg d$}, or,
equivalently,
\mbox{$p \wedge q \wedge \neg r \vee \neg p \wedge \neg q \wedge r$}.
The common wisdom, defended in~\cite{Gin:86}, would like to convince us that 
we should not presume $q$ to be true.
The position defended here, presumes that
$q$ is true (and also $p$ and $\neg r$) because this situation violates
only one default $( : r )$ whereas the other possible situation:
\mbox{$\neg p \wedge \neg q \wedge r$} violates two defaults.
}
\end{example}

The reader may suspect that our policy gives results that are extremely 
sensitive
to the way the defaults are presented.
Indeed, the way defaults are presented is important,
and our perspective on defaults does not enjoy the nice global properties
of rational closure described in~\cite[Section 5.5]{LMAI:92} that make
it invariant under the addition or deletion of entailed defaults.
Two examples of this phenomenon will be described now.
The first one shows that the addition to $D$ of a default entailed by $D$ may add new
conclusions.
The second one shows that the addition to $D$ of a default entailed by $D$ may 
force us to withdraw previous conclusions.
The examples presented are very simple and natural and should convince the
reader that any presumptive reading of defaults leads to a high
sensitivity to the presentation of the default information.
This sensitivity is, probably, a drawback of the lexicographic closure.
The following examples should convince the reader that this problem is 
inevitably brought about by a presumptive understanding of defaults.
If we had decided to consider multisets of defaults, instead of sets,
thus allowing certain (stronger) defaults to appear a number of times in $D$,
our construction would had been sensitive to the number of times each default
appears in $D$.
\begin{example}[Adding entailed defaults may add conclusions]
{\rm
Let $D$ be the singleton \mbox{$\{( : p \wedge q)\}$}.
The default (identifying defaults and conditional assertions)
\mbox{$( : p )$} is obviously entailed by $D$.
But the default \mbox{$(\neg p \vee \neg q : p)$} is not entailed
by $D$, the antecedent being inconsistent with the only default of $D$.
Nevertheless \mbox{$(\neg p \vee \neg q : p)$} is entailed by the set
\mbox{$\{ ( : p \wedge q ) , ( : p ) \}$}, since its antecedent
is consistent with the second default.
The behavior of the corresponding Poole system is the same.
}
\end{example}

\begin{example}[Adding entailed defaults may delete conclusions]
{\rm
Let $D$ be the set \mbox{$\{ ( : p ) , ( : q ) \}$}.
Both defaults \mbox{$( : p \leftrightarrow q )$} and
\mbox{$( \neg p : q )$} are entailed by $D$.
But \mbox{$( \neg p : q )$} is {\em not} entailed by the set
\mbox{$\{ ( : p ) , ( : q ) , ( : p \leftrightarrow q )\}$},
since the antecedent is consistent with both the last defaults separately
but not together, and there are therefore two maxbases.
In this case also, the behavior of the corresponding Poole systems is the same.
}
\end{example}

\section{Lexicographic closure}
\label{sec:full}
\subsection{Introduction and definition}
\label{subsec:introfull}
In the previous Section we discussed the treatment of conflicting
defaults that had the same precedence.
We shall now treat arbitrary conflicting defaults, and define the construction
we propose in full generality.
We must take into account the fact that defaults may have
different weight, or precedence. Fortunately, the correct definition of the
relative precedence of defaults has been obtained in a previous work.
Given a finite set of defaults $D$, the precedence of a default is given by 
its rank (higher rank means higher precedence), i.e., by the rank of 
its antecedent
as defined in~\cite[Section 2.6]{LMAI:92}.
The definition presented here is equivalent to the original
definition, by Corollary 5.22 there.

We shall now remind the reader of this definition.
Let $D$ be a finite set of defaults, and $\tilde{D}$ the set of its material
counterparts. Let $a$ be a formula.
We shall put \mbox{$E_{0} = D$}.
If $a$ does not have rank less than $i$, 
but is consistent with $\tilde{E_{i}}$,
it has rank $i$.
The set $E_{i + 1}$ is the subset of $E_{i}$ that contains all defaults 
\mbox{$( a : b)$} of $D$ for which $a$ does not have rank less or equal
to $i$.
We shall put \mbox{$D_{i} = E_{i} - E_{i + 1}$}, and let $D_{\infty}$
be the set of all elements of $D$ that have no rank, i.e., have infinite rank.
Elements of $D_{\infty}$ have precedence over all other defaults.
Notice that, since $D$ is finite, 
all $D_{i}$'s are empty after a certain point, except possibly
$D_{\infty}$.
There is a $k$ such that for any $i$, \mbox{$k \leq i < \infty$}, 
\mbox{$D_{i} = \emptyset$}. The smallest such number $k$ will be called the 
{\em order} of the set $D$.
The set $D$ may be partitioned into 
\mbox{$D_{\infty} \oplus D_{k-1} \oplus D_{k-2} \oplus \ldots \oplus D_{0}$}.

We remind the reader that the rational closure of the set $D$,
defined in~\cite{Leh:89} and studied in depth in~\cite{LMAI:92} 
(see Theorem 5.17 and Lemma 2.24) is the set of
defaults $\overline{D}$ that consists of all defaults \mbox{$(a : b )$}
such that the rank of $a$ is strictly less than the rank of 
\mbox{$a \wedge \neg b$} (equivalently, the rank of \mbox{$a \wedge b$} 
is strictly less than the rank of \mbox{$a \wedge \neg b$}), or such that $a$
has no rank. We shall now define another closure for $D$, the lexicographic 
closure. We define the lexicographic closure by way of a modular model
in which every model is ranked by the set of defaults it violates.
A similar presentation may be used to define rational closure,
it will also be described.

\subsection{The model-theoretic description}
As usual, we shall suppose a finite set $D$ of defaults is given.
We shall describe the consequence relation defined by $D$,
the lexicographic closure of $D$, $D^{l}$ as the consequence relation
defined by a certain modular model, $\cM_{D}$.
To define this model, we need to order the propositional models
by some modular ordering. We shall order the propositional models by ordering
the sets of defaults (of $D$) that they violate:
each model $m$ violates a set $D_{m} \subseteq D$ of defaults.
How should we order the subsets of $D$?
Intuitively, we are looking for a ``degree of seriousness''.
We prefer to violate a ``lighter'' set of defaults than a more
serious one, i.e., a propositional model that violates a lighter
set of defaults is more normal than a model that violates a more serious set.
There are two criteria that must be taken into account when
deciding which of two sets is more serious:
\begin{itemize}
\item the size of the set: the smaller the set, the less serious it is.
We have seen in Section~\ref{sec:Dsys}, that ``smaller'' should be taken
here to mean ``of smaller size'', and
\item the seriousness of the elements of the set, it is less serious
to violate a less specific default than a more specific default,
i.e. a default of lower rank than a default of higher rank. 
\end{itemize}
The reader should notice here that our definition is in no way circular.
The lexicographic closure is defined in terms of a specific modular model
that is, in turn, defined in terms of the ranks of the formulas involved.
These ranks have been defined above, by a straightforward inductive definition.
The ranks of the formulas have a close relationship with the ordering
of the modular model that defines the rational closure of $D$, but this is 
a different modular model. In fact, the model we are describing now
is a refinement of the model that defines rational closure
(a level may split into a number of sublevels).
The next question, now, is how should we compose those two criteria?
The principle of rationality will trace the way for us.
We want a modular ordering on the subsets of $D$.
Each one of the criteria above gives a modular ordering.
Is there a general way to combine two modular orderings and obtain
a modular ordering? Yes, a lexicographic (i.e. consider one criterion as
the principal criterion, the other as secondary) composition of modular
orderings is a modular ordering.
Which of the two criteria above should be considered as the major
criterion?
Clearly the second one: specificity. We should prefer violating two
defaults of low specificity to violating one of high specificity.
\begin{example}
{\rm
Let \mbox{$D = \{ (:p) , (:q) , (:x) , (y:\neg x) , (y:r) \}$}.
Suppose our assumptions are 
\mbox{$ y \wedge (\neg p \wedge \neg q \vee \neg r)$}.
You may imagine that $p$, $q$ and $x$ are generic properties (of birds, say),
and that $y$ is a class of birds that are exceptional with respect to $x$.
The property $r$ is a generic property of $y$ birds.
Suppose we have a bird that is part of the $y$ class, and is known to be,
either exceptional with respect to two generic properties of birds,
or exceptional with respect to one generic property of the sub-class $y$.
Presumption of typicality (see Section~\ref{subsec:rat}), from the
last default of $D$ proposes the conclusion $r$ (and therefore
$\neg p$ and $\neg q$).
Presumption of independence proposes the conclusions $p$ and $q$ 
(and therefore $\neg r$).
Priority to typicality convinces us to accept
the former and reject the latter.
}
\end{example}
Therefore, to decide which of two sets of defaults is more serious,
we shall partition those sets into subsets of defaults of equal ranks
and compare (by size) rank by rank, starting with the higher ranks.
As soon (in terms of ranks) as a decision can be made, we decide and stop.
\begin{definition}
\label{def:tnumber}
Let $D$ be a set of defaults and $k$ its order.
To every subset \mbox{$X \subseteq D$} may be associated a $k + 1$-tuple
of natural numbers: \mbox{$\langle n_{0} , \ldots , n_{k} \rangle$},
where \mbox{$n_{0} = \mid D_{\infty} \cap X \mid$},
\mbox{$n_{1} = \mid D_{k - 1} \cap X \mid$},
and in general, for \mbox{$i = 1 , \ldots , k$}, 
\mbox{$n_{i} = \mid D_{k - i} \cap X \mid$}.
In other terms, $n_{0}$ is the number of defaults of $X$ that have no rank
and, for \mbox{$0 < i \leq k$},
$n_{i}$ is the number of defaults of $X$ that have rank $k-i$.
We shall order the subsets of $D$ by the natural lexicographic ordering
on their associated tuples. This is a strict modular partial ordering:
it will be denoted by $\prec$, (the {\em seriousness ordering}).
\end{definition}
The seriousness ordering on sets of defaults is used to order
the propositional models: \mbox{$m \prec m'$} iff \mbox{$V(m) \prec V(m')$},
where $V(m) \subseteq D$ is the set of defaults violated by $m$.
This modular ordering on models defines a modular preferential model,
that, in turn defines a consequence relation, $D^{l}$, 
the lexicographic closure of $D$.
The reader may check that all examples treated in this paper conform to
the definition above.

Let us, now, before we give an alternative description of lexicographic 
closure, briefly digress to see that rational closure may be defined
by a specific seriousness ordering, different from the one defined in
Definition~\ref{def:tnumber}.
\begin{definition}
\label{def:tnumber2}
Let $X_{1}$ and $X_{2}$ be subsets of a set $D$ of defaults and $k$ its order.
Let $n^{1}_i$ and $n^{2}_i$, for \mbox{$i = 1 , \ldots , k$},
be the size of the partitions of $X_{1}$ and $X_{2}$ respectively.
Let $m^{j}$ be the smallest $i$ such that \mbox{$n^{j}_{i} \neq 0$},
for \mbox{$j = 1 , 2$}.
We shall write \mbox{$X_{1} \ll X_{2}$} iff \mbox{$m^{1} > m^{2}$}.
\end{definition}
Clearly \mbox{$X_{1} \ll X_{2}$} implies \mbox{$X_{1} \prec X_{2}$},
i.e., $\ll$ is coarser than $\prec$.
\begin{theorem}
\label{the:ll}
Suppose $a$ has a finite rank. 
The conditional assertion \mbox{$a \NIm b$} is a member of
the rational closure of $D$ iff it is satisfied by the modular model in which
each propositional model is ranked by the $\ll$ ordering on the set of
defaults it violates.
\end{theorem}
\proof
Suppose the rank of $a$ is strictly less than that of \mbox{$a \wedge \neg b$},
and that $m$ is a propositional model that satisfies $a$ and is minimal
among those for the $\ll$ ordering. If $a$ has rank $l$, there is a model
that satisfies $a$ and violates no default of $D$ of rank greater or equal to
$l$. We conclude that $m$ violates no such default and therefore satisfies 
no formula of rank strictly greater than $l$.
The model $m$ does not
satisfy \mbox{$a \wedge \neg b$}, and therefore satisfies $b$.

Suppose, now, that all propositional models that satisfy $a$ and are
minimal in the $\ll$ ordering for that property, also satisfy $b$.
Let the rank of $a$ be $k$. Since there is a propositional model
that satisfies $a$ and violates no default of rank greater or equal to
$k$, all models that satisfy $a$ and violate no default of rank greater
or equal to $k$ satisfy $b$. We conclude that the rank of 
\mbox{$a \wedge \neg b$} is greater than $k$.
\QED
We may now show that the lexicographic closure is a superset of the rational
closure, at least for defaults of finite rank, thus almost complying with
Thesis~5.25 of~\cite{LMAI:92}.
\begin{theorem}
\label{the:Thesis5.2}
If $a$ has a finite rank and \mbox{$a \NIm b$} is an element of the rational
closure of $D$, then it is an element of its lexicographic closure.
\end{theorem}
\proof
Suppose $a$ has a finite rank and \mbox{$a \NIm b$} is an element of 
the rational closure of $D$.
By Theorem~\ref{the:ll}, \mbox{$a \NIm b$} is satisfied in the modular
model defined by $\ll$. It is therefore satisfied in any modular model
defined by a finer relation. We noticed, just following 
Definition~\ref{def:tnumber2}, that $\prec$ is such a finer relation.
\QED
A characterization of the lexicographic closure in terms of bases will be 
described now.
\subsection{Bases}
\begin{definition}
\label{def:base}
Let $a$ be a formula, and $B$ a subset of $D$. We shall say that
$B$ is a {\em basis} for $a$ iff $a$ is consistent with $\tilde{B}$,
the material counterpart of $B$, and, $B$ is maximal {\em with respect to
the seriousness ordering} for this property.
\end{definition}
The following lemma may help explain the structure of bases, but is not
used in the sequel.
\begin{lemma}
\label{le:bases}
If $a$ has rank $i$ ($a$ has no rank is understood as $a$ 
having an infinite rank)
and $B$ is a basis for $a$, then, for any 
\mbox{$j \geq i$}, \mbox{$D_{j} \subseteq B$}.
\end{lemma}
In other terms, any basis for $a$ is full, for all indexes larger or equal
to the rank of $a$.
\proof
Since $a$ has rank $i$, for any \mbox{$j \geq i$}, $a$
is consistent with \mbox{$E_{j}$} and therefore with
\mbox{$ D_{j} \cup B \cap E_{j + 1} \subseteq E_{j}$}.
\QED
\begin{theorem}
\label{the:entail}
The default \mbox{$(a : b)$} is in $D^{l}$, 
the lexicographic closure $D$ iff,
for any basis $B$ for $a$, \mbox{$\tilde{B} , a \models b$}.
\end{theorem}
\proof
The proof is a generalization of that of Theorem~\ref{the:rat}.
Let $k$ be the order of $D$.
Let \mbox{$d_{i} = \mid D_{i} \mid$} for 
\mbox{$i = 0 , \ldots , k-1 , \infty$}.
\begin{enumerate}
\item \label{aaa}
If $a$ is satisfied by some model of seriousness level
\mbox{$(i_{0} , \ldots , i_{k})$},
then, there is a basis for $a$ of level
\mbox{$(d_{\infty} - i_{0} , d_{k-1} - i_{1} , \ldots , d_{0} - i_{k})$},
and all bases for $a$ have this level.
\item \label{bbb}
If $B$ is a basis of level \mbox{$(l_{0} , \ldots , \l_{k})$} for $a$,
then, there is a model of level 
\mbox{$(d_{\infty} - l_{0} , d_{k-1} - l_{1} , \ldots , d_{0} - l_{k})$}
that satisfy $a$, and no model of strictly smaller seriousness satisfy $a$.
\item \label{ccc}
There is no basis for $a$ iff $a$ is a logical contradiction.
\end{enumerate}

Suppose, first, that \mbox{$a \NIm b \in D^{l}$}.
If no model satisfies $a$, $a$ is a logical contradiction and
\mbox{$ X , a \models b$} for any $X$.
Suppose, then, that $a$ is satisfied by some model of level 
\mbox{$(i_{0} , \ldots , i_{k})$}, 
but by no model of lower level. Any model of level 
\mbox{$(i_{0} , \ldots , i_{k})$} that satisfies
$a$, satisfies $b$, by hypothesis. 
Let $B$ be a basis for $a$, of seriousness 
\mbox{$(b_{\infty} , \ldots , b_{0})$}. 
By remark~(\ref{bb}), \mbox{$ d_{j} \: - b_{k-j} \: = \: i_{k-j}$},
for \mbox{$j = 0 , \ldots , k-1$}, and
\mbox{$d_{\infty} - b_{0} = i_{0}$}.
Any model that satisfies $B$ is obviously of seriousness level 
less or equal to
\mbox{$(d_{\infty} - b_{0} ,  d_{k-1} - b_{1} , \ldots , d_{0} - b_{k})$},
i.e., of level less or equal 
\mbox{$i_{0} , \ldots , i_{k}$}.
Any model that satisfies $B$ and $a$ is therefore of level 
\mbox{$i_{0} , \ldots , i_{k}$},
and satisfies $b$, by hypothesis. 
We conclude that \mbox{$B , a \models b$}.

Suppose, now, that for any basis $B$ for $a$ we have
\mbox{$B , a \models b$}.
If there is no basis for $a$, then, by remark~(\ref{cc}), 
$a$ is a logical contradiction and \mbox{$a \NIm b \in D^{l}$}.
Suppose, then, the bases for $a$ are of seriousness
\mbox{$(b_{0} , \ldots , b_{k})$}.
There is, by remark~(\ref{bb}),  
a model of $a$ of level 
\mbox{$l = (d_{\infty} - b_{0} , \ldots , d_{0} - b_{k})$}, 
and there is no model of level less than $l$ that satisfies $a$.
We must show that any model of $a$ of level $l$ satisfies $b$.
Let $m$ be such a model. Since $m$ violates $d_{j} - b_{k-j}$ defaults
of rank $j$, 
it satisfies a set $M$ containing $b_{k-j}$ defaults of rank $j$. 
But $M$ is consistent with $a$,
since $m \models a$. The seriousness of $M$ is 
\mbox{$(b_{0} , \ldots , b_{k})$}, the seriousness
of the bases for $a$.
Therefore $M$ is a basis for $a$ and,
since \mbox{$M , a \models b$}, we conclude that \mbox{$m \models b$}.
 
\QED
We shall now describe the lexicographic closure of a number
of sets of defaults, some of them well-known from the literature.
\section{Examples}
\label{sec:exc}
%Exceptions. More exceptional defaults have precedence.

In this Section, motivating examples will be described, indicating,
for each of them the conclusions endorsed by the lexicographic closure.
My goal is to, gradually, convince the reader, that each one of the 
decisions taken in the process of defining the lexicographic closure was
reasonable.
My goal is {\em not} to convince the reader that lexicographic
closure provides the {\em intuitively} correct answer once the propositional
variables have been interpreted in some manner that is well-known in the 
folklore of the field, because I believe that, in most cases, we intuitively
treat interpreted formulas in a meaning-dependent manner, not in the formal,
meaning-independent way that is the hallmark of logic.
In other terms, once the variables are interpreted, there is no way of
knowing whether the intuitive conclusions come from formal logical 
considerations or from world knowledge the reasoner has about the situations
or objects the interpreted variables refer to.
A first example exemplifies why one needs to give precedence to defaults
describing exceptional cases over those that describe more normal cases.
\begin{example}[Exceptions]
{\rm
Let \mbox{$D = \{ ( : p ) , ( : q ) , ( \neg p : \neg q )\}$}.
Here $p$ and $q$ hold by default, and when $p$ does not hold, then,
by default $q$ does not hold either.
Suppose we know that $p$ does not hold. Then, obviously we cannot
use the first default. But, we could use the second one to conclude $q$
or the third one to conclude $\neg q$. We obviously want to presume
$\neg q$, and we need to say that the third default has precedence
over the second one. Fortunately, it is not difficult to justify
why the third default has precedence over the other ones.
It (i.e., its antecedent) 
has rank one
whereas the two other defaults have rank zero.
This comes from the fact that the antecedents of the first two defaults
({\bf true}), describe some unexceptional situation, while $\neg p$
describes an unexpected, exceptional situation since $p$ is, by default,
presumed to be true.

The technical description of the lexicographic closure follows.
The order of $D$ is two.
The first two defaults of $D$ have rank zero, the last one has rank one.
The most normal models (those that have level zero), are those that
satisfy $p$ and $q$ (and therefore \mbox{$\neg p \ra \neg q$}).
On level one, we find those models that satisfy \mbox{$\neg p \ra \neg q$}
and violate exactly one of $p$ or $q$: the models that satisfy 
$p$ and $\neg q$.
The third level contains those models that violate both $p$ and $q$,
but satisfy \mbox{$\neg p \ra \neg q$}, i.e. the models satisfying
$\neg p$ and $\neg q$.
The fourth level contains all models violating \mbox{$\neg p \ra \neg q$},
but satisfying $p$ and $q$; it is empty.
The fifth level contains all models violating \mbox{$\neg p \ra \neg q$},
and exactly one of $p$ and $q$: it contains one model.
The sixth level is empty.
}
\end{example}
So, we must give precedence to defaults of higher rank over defaults of
lower rank.
Notice that rank is really all the difference between the default
\mbox{$( a : b )$} and the default \mbox{$( : a \ra b )$}.
The second one has always rank zero, while the first one may have a much higher
rank (if $a$ is presumed to be false) and is therefore more powerful.
The reader may easily check that the first two defaults of $D$ have rank
zero, whereas the third one has rank one.
Let us now treat the similar but more classical penguin example.
\begin{example}[Penguins]
{\rm
Let \mbox{$D = \{ ( p : q ) , (r : p) , (r : \neg q ) \}$}.
The default \mbox{$( : \neg r )$} is entailed by $D$ and
the second and third defaults have rank one, 
whereas the first default has
rank zero.
The defaults \mbox{$( r : \neg q )$} and \mbox{$( p \wedge r : \neg q )$}
are entailed by $D$ whereas \mbox{$( r : q )$} is not.

Technically, the rank of $D$ is two.
At level zero: all models satisfying $p \ra q$ and $\neg r$.
At level one: all models satisfying $p$ and $\neq q$, and those 
satisfying $r$, $\neg p$ and $\neg q$.
Level two is empty.
Level three: all models satisfying $r$, $q$ and $p$.
Level four: all models satisfying $r$, $q$ and $\neg p$.
Level five: empty.
}
\end{example}
In the present proposal, 
this precedence of defaults of higher rank is in a sense 
(or in two ways) absolute: one should not trade the violation of a default
of rank $n + 1$ for the violation of any number of defaults of rank less than
or equal to $n$. On this point, the present proposal is in disagreement with
the Principle of Maximal Entropy as proposed in~\cite{GMP:93}.
The following example will exhibit this disagreement, it is not meant
to support one construction against the other.
\begin{example}[Winged Penguins]
{\rm
Let \mbox{$D = \{ ( b : w ) , ( b : f ) , ( p : b ) , ( p : \neg f ) \}$}.
The default \mbox{$( p \wedge ( f \vee \neg w ) : b )$} is entailed
by $D$, since the only basis for \mbox{$ p \wedge ( f \vee \neg w )$} is
the set \mbox{$\{( p : b ) , ( p : \neg f )\}$}, 
containing all defaults of rank one.
In fact, even 
\mbox{$( p \wedge ( f \vee \neg w ) : b \wedge \neg f \wedge \neg w)$}
is entailed by $D$.
But the Principle of Maximal Entropy of~\cite{GMP:93} will consider 
as equivalent
\begin{itemize}
\item
to violate two defaults of rank zero: 
(\mbox{$ ( b : w )$} and \mbox{$ ( b : f )$},
and 
\item to violate one default of rank one: \mbox{$( p : b )$}, 
\end{itemize}
and therefore
will not accept \mbox{$( p \wedge ( f \vee \neg w ) : b )$}.
Notice that the set $D$ is Minimal Core in the sense of~\cite{GMP:93}.
}
\end{example}
\begin{example}[Exceptions again]
{\rm
Let 
\[
D = \{ ( : r ) , ( : p ) , ( : q ) , ( \neg p : \neg q ), 
( \neg p : \neg r ) \}.
\]
Suppose our specific information is \mbox{$\neg p \wedge q$},
which means the situation is doubly exceptional: $p$ is presumed true
but is in fact false, and when $p$ is false $q$ is presumed false, but
it is true. In other words the rank of \mbox{$\neg p \wedge q$} is two.
Should we presume $r$ to be true or false? 
It is clear we should presume it false, since the default 
\mbox{$( \neg p : \neg r )$} talks about a situation closer to the
one at hand than the default \mbox{$( : r )$}, and should have
precedence over it. But notice that our information shows that
the situation described by \mbox{$\neg p \wedge q$}
is exceptional with respect to the one described by
\mbox{$ \neg p $}.
}
\end{example}

\section{Discussion}
\label{sec:properties}
The lexicographic closure $D^{l}$ of a finite set $D$ of defaults
is defined by a modular model in which all propositional models
appear, at some level.
An assertion of the form \mbox{$a \NIm {\bf false}$} will appear in
$D^{l}$ only if $a$ is a logical contradition.
In other terms, lexicographic closure is, in the terminology 
of~\cite{Mak:Handbook},
{\em consistency preserving}. This is indeed one of the hallmarks
of Default Reasoning {\em \`{a} la Reiter} 
as attested by the discussion in~\cite{Reiter:80}, 
and in particular Corollary~2.2 there.

Since there has been a lot of discussion in the literature, in particular
in~\cite{Gins:87} and \cite{FLMo:91}, of the principle of Transitivity,
it is probably worth a short discussion.

The question of Transitivity is: should we accept \mbox{$(a : c)$}
on the basis of the two defaults  \mbox{$(a : b)$} and \mbox{$(b : c)$}?
The answer proposed here, and which follows from our construction, 
is that, if we have both 
\mbox{$(a : b)$} and \mbox{$(b : c)$}, and if $a$ and $b$ are both of the same
rank, then we should also accept \mbox{$(a : c)$}.
Note that, if we accept \mbox{$(a : b)$}, then $b$ has rank lower or equal
to that of $a$: indeed if the rank of $b$ was larger than that of $a$, 
the rational closure of our set of defaults would include
\mbox{$a \vee b \NIm \neg b$} and therefore \mbox{$a \NIm \neg b$},
and, by Theorem~\ref{the:Thesis5.2}, we would accept \mbox{$( a : \neg b )$}
in the lexicographic closure.
If the rank of $a$ is strictly greater than that of $b$,
then we are not guaranteed that\mbox{$(b : c)$} will be part of all bases for
$a$.

Given a finite set $D$ of defaults and a default \mbox{$(a : b)$}, 
how difficult is it to decide whether \mbox{$(a : b)$} is entailed by $D$?
This decision seems to require the computation of the ranks of the defaults
of $D$, but this is relatively easy: a quadratic number of satisfiabilty
problems.
It seems that it also requires the consideration of a possibly
large number of subsets of $D$ and seems therefore inherently exponential.
The rational closure construction of~\cite{LMAI:92} provides a quick
and dirty approximation to this construction in the following sense:
if a default belongs to the rational closure it is entailed
(up to a slightly different treatment of formulas that have no rank).
The case in which all defaults have a Horn structure, needs further study.
One may perhaps avoid the exponential blow-up in this case.

\section{Acknowledgements}
Comments, corrections and suggestions by 
Michael Freund, Mois\'{e}s Goldszmidt and David Makinson 
are gratefully acknowledged.
Three anonymous referees (not necessarily disjoint from the previous set)
also helped significantly to make this paper more readable.

\end{document}